\documentclass[runningheads]{llncs}
\usepackage{times}
\usepackage{latexsym}
\usepackage{graphicx}
\usepackage{amssymb}
\usepackage{amsmath}
\usepackage{multirow}

\usepackage{url}
\usepackage{xspace}
\usepackage{sidecap}

\newcommand{\diagram}{\textsc{Diagram}\xspace}
\newcommand{\dnode}{\textsc{Diagram}$_{node}$\xspace}
\newcommand{\dedge}{\textsc{Diagram}$_{edge}$\xspace}

\newcommand{\hope}{\textsc{hope}\xspace}
\newcommand{\deepwalk}{\textsc{Deepwalk}\xspace}
\newcommand{\ntov}{\textsc{Node2Vec}\xspace}
\newcommand{\vgae}{\textsc{vgae}\xspace}
\newcommand{\dane}{\textsc{dane}\xspace}
\newcommand{\anrl}{\textsc{anrl}\xspace}

\newcommand{\sg}{\textsc{SkipGram}\xspace}

\author{Zekarias T. Kefato\inst{1} \and
Nasrullah Sheikh\inst{2}\and
Alberto Montresor\inst{2}}
\authorrunning{Z. T. Kefato et al.}
\institute{Royal Institute of Technology, Stockholm, Sweden\\
\email{zekarias@kth.se}\\
 \and
University of Trento\\
\email{\{nasrullah.sheikh,alberto.montresor\}@unitn.it}}

\title{Which way?\\
Direction-Aware Attributed Graph Embedding\thanks{Presented at the GEM: Graph Embedding and Mining Workshop collocated with ECML-PKDD 2019 Conference. The source code is provided in the following github repo: https://github.com/zekarias-tilahun/diagram} }

\date{}

\begin{document}
\maketitle
\begin{abstract}
  Graph embedding algorithms are used to efficiently represent (encode) a graph in a low-dimensional continuous vector space that preserves the most important properties of the graph.
  One aspect that is often overlooked is whether the graph is directed or not. 
  Most studies ignore the directionality, so as to learn high-quality representations optimized for node classification.
  On the other hand, studies that capture directionality are usually effective on link prediction but do not perform well on other tasks.

  This preliminary study presents a novel text-enriched, direction-aware algorithm called \diagram, based on a carefully designed multi-objective model to learn embeddings that preserve the direction of edges, textual features and graph context of nodes.
  As a result, our algorithm does not have to trade one property for another and jointly learns high-quality representations for multiple network analysis tasks.
  We empirically show that \diagram significantly outperforms six state-of-the-art baselines, both direction-aware and oblivious ones, on link prediction and network reconstruction experiments using two popular datasets.
  It also achieves a comparable performance on node classification experiments against these baselines using the same datasets.
\end{abstract}

\section{Introduction}
\label{sec:introduction}

Recently, we have seen a tremendous progress in graph embedding algorithms that are capable to learn high-quality vector space embeddings of graphs.
Most of them are based on different kinds of neural networks, such as the standard multi-layer perceptron (MLP) or convolutional neural networks (CNN).
Regardless of the architecture, most of them are optimized to learn embeddings  for effective node classification.

\begin{figure}[t!]
\centering
\includegraphics[scale=0.6]{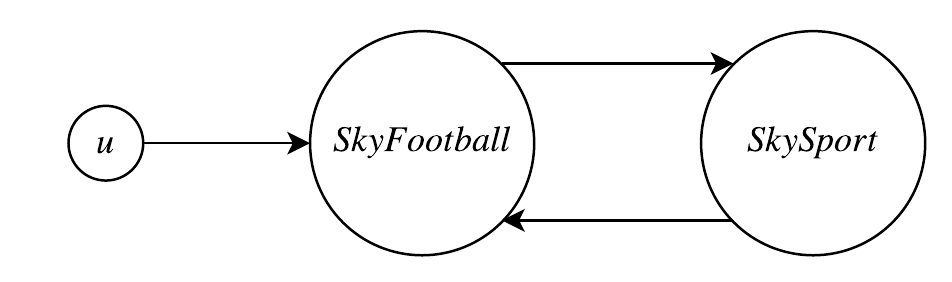}
\caption{An example follower graph between a random football enthusiast Twitter user and two popular football and sport news channels Twitter accounts.}
\label{fig:transitivity_in_graphs}
\end{figure}

Even though they have been shown to be useful for link prediction, they are only capable to predict if two nodes are connected; they are not able to identify the direction of the edge.
For example, in social networks like Twitter, it is very likely for ``organic users'' (such as $u$ in Fig.~\ref{fig:transitivity_in_graphs})  to follow popular users like \emph{SkyFootball}, based on their interest on the provided content, while the follow-back is very unlikely.
Consequently, transitivity from $u$ to $\mathit{SkySport}$ is usually asymmetric as empirically demonstrated by~\cite{Ou:2016:ATP:2939672.2939751}.
For this reason, a good embedding algorithm over directed graphs should not be oblivious to directionality.

Another pressing issue in graph embedding research is the need for preserving similarity (homophily) between nodes.
For example, the similarity between $u$ and \emph{SkyFootball} is usually dependent on the undirected neighborhood and the content produced/consumed by them~\cite{Pan:2016:TDN:3060832.3060886,ijcai2018-467}.
Thus, it is more important to examine the content and the entire neighborhood of nodes to measure their similarity.

Existing studies usually trade-off between preserving directionality versus similarity~\cite{Ou:2016:ATP:2939672.2939751,ijcai2018-467,Wang:2016:SDN:2939672.2939753,ijcai2018-438,Perozzi:2014:DOL:2623330.2623732,Grover:2016:NSF:2939672.2939754,DBLP:journals/corr/TangQWZYM15,Yang:2015:NRL:2832415.2832542,Yang:2017,Hamilton:2017,Kipf:2016,Pan:2016:TDN:3060832.3060886,gaticlr2018,DBLP:journals/corr/abs-1806-01973}.
Hence, they either learn high-quality embeddings for classification at the expense of asymmetric link prediction, or vice versa.

In this study, we argue that we do not have to choose between these aspects and we propose a novel algorithm called \diagram (\textbf{Di}rection-Aware \textbf{A}ttributed \textbf{Gr}aph E\textbf{m}bedding) that seeks to learn embeddings that balance the two aspects: 
\diagram preserves directionality and hence asymmetric transitivity, while at the same time striving to preserve similarity.
We achieve this by designing a multi-objective loss function that seeks to optimize with respect to (1) direction of edges (2) content of nodes (Binary or TF-IDF features) and (3) undirected graph context.

First we propose a variant of \diagram called \dnode; after investigating its limits, we propose an extension called \dedge to address them.

The paper is structured as follows.
In Section~\ref{sec:model} we give a description of the proposed model and report the empirical evaluation in Section~\ref{sec:experiments}.
In Section~\ref{sec:related_work} we discuss related work.  Section~\ref{sec:conclusion} concludes the paper.

\section{Model}
\label{sec:model}

Let $G = (N, E, \mathbf{M}, \mathbf{D})$ be a directed graph, where $N$ is a set of $n$ nodes and $E$ is a set of $m$ edges.
$\mathbf{M} \in \mathbb{R}^{n \times n}$ is an adjacency matrix, binary or real, and $\mathbf{D} \in \mathbb{R}^{n \times d}$ is a feature matrix, for instance constructed from a document corpus $\mathcal{D} = \lbrace D_1, \ldots, D_n \rbrace $, where $D_u$ is a document associated to node $u \in N$.
Each word $w \in D_u$ is a sample from a vocabulary $V$ of $d$ words.
Matrix $\mathbf{D}$ can be a simple binary matrix, where $\mathbf{D_{u, i}}=1$ means that the word, the $i$-th word of $V$, appears in document $D_u$ of user $u$.
It can also be a TF-IDF weight that captures the importance of word $w$ to node $u$.


For every node $u \in N$, our objective is to identify three embedding vectors, $\mathbf{e_u} = \lbrace \mathbf{z_u}, \mathbf{o_u}, \mathbf{i_u} \rbrace$.
A recent study~\cite{Epasto:2019:SEE:3308558.3313660} has shown the power of learning multiple embedding vectors of each node that capture different social contexts (communities).
In our study, however, each embedding vector is designed to meet the three goals that we set out to achieve in the introduction, which are:
\begin{enumerate}
\item Undirected neighborhood and content similarity-preserving embedding -- $\mathbf{z_u} \in \mathbb{R}^k, k \ll n$ -- based on $\mathbf{M_u} | \mathbf{M^T_u}$ and $\mathbf{D}$, where $|$ is a bitwise or.
\item Outgoing neighborhood-preserving embedding -- $\mathbf{o_u} \in \mathbb{R}^k$ -- based on $\mathbf{M_u}$
\item Incoming neighborhood-preserving embedding -- $\mathbf{i_u} \in \mathbb{R}^k$ -- based on $\mathbf{M^T_u}$
\end{enumerate}

To motivate our work, suppose we want to estimate the proximity between $u$ and $v$. 
If $(u, v) \in E$ is a reciprocal edge, then the estimated proximity should be equally high for both $(u, v)$ and $(v, u)$, otherwise it should be high for $(u, v)$ and very small for $(v, u)$. 

To this end, we adopt the technique proposed by~\cite{Ou:2016:ATP:2939672.2939751} for estimating the asymmetric proximity between $u$ and $v$ as the dot product between $\mathbf{o_u} \in \mathbb{R}^k$ and $\mathbf{i_v} \in \mathbb{R}^k$ as:
\[
\texttt{prx}(u, v) \propto \texttt{dot}(\mathbf{o_u}, \mathbf{i_v})
\] 
which are the learned embeddings of node $u$ and $v$ that capture their outgoing and incoming neighborhood, respectively.
\begin{remark}\label{rem:proximity_remark}
To satisfy the above condition, if there is an unreciprocated directed edge $(u, v) \in E$ we consider that as a high non-symmetric proximity, and hence $\mathbf{o_u}$ and $\mathbf{i_v}$ should be embedded close to each other.
\end{remark}

Thus, our goal is to jointly learn three embeddings $\mathbf{e_u} = \lbrace \mathbf{z_u}, \mathbf{o_u}, \mathbf{i_u} \rbrace$ of each node $u \in N$ using a single shared autoencoder model as follows.

\begin{figure}[t!]
\centering
\includegraphics[scale=0.55]{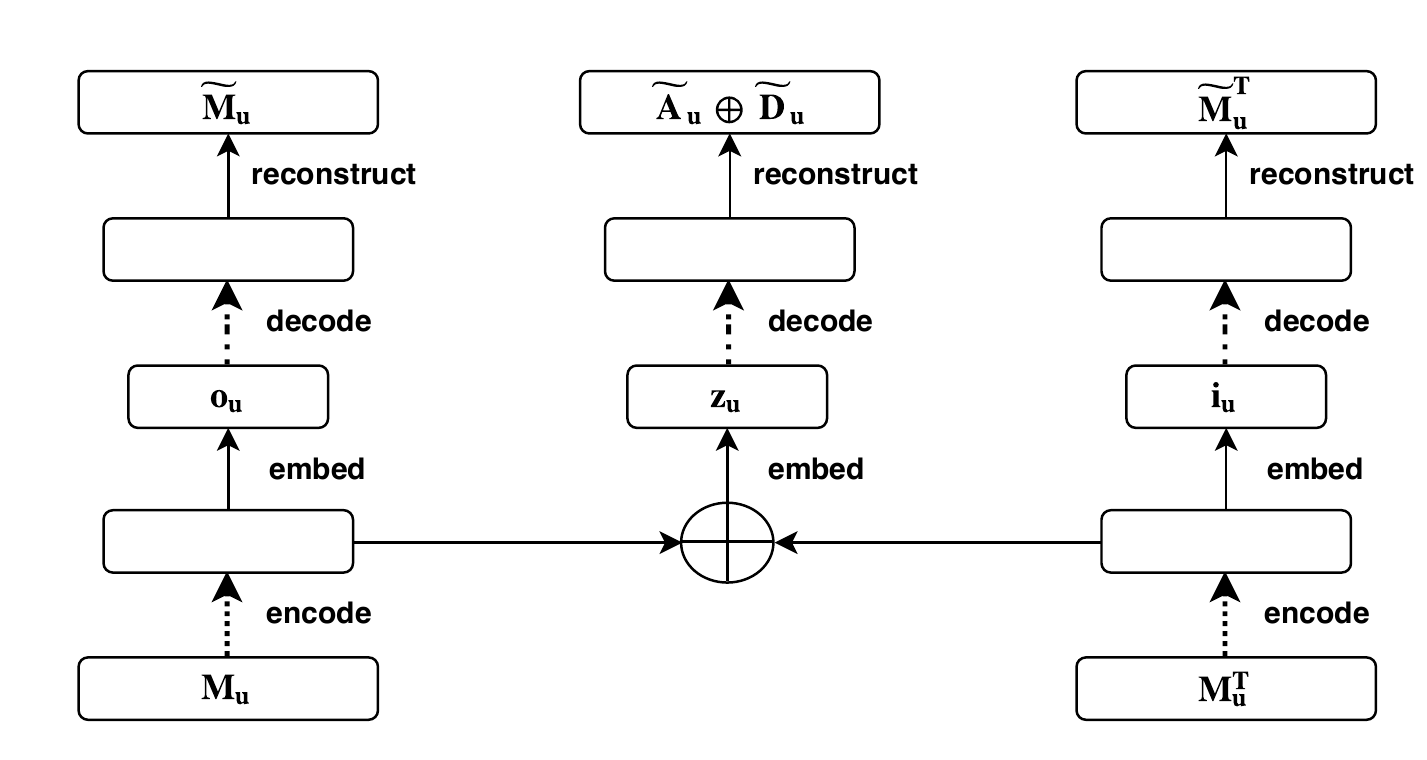}
\caption{The \diagram architecture. Dotted and solid arrows indicate an MLP and SLP, respectively}
\label{fig:diagram}
\end{figure}

\subsection{Node Model} 
The first variant of \diagram is \dnode (Fig.~\ref{fig:diagram}); for each node $u$, its inputs are $\mathbf{M_u}$ and $\mathbf{M^T_u}$.
It has four components, which are \texttt{encode}, \texttt{embed}, \texttt{decode}, and \texttt{reconstruct}.
The \texttt{encode} and \texttt{decode} components are formulated as a multi-layer perceptron (MLP), with $L_{enc}$, and $L_{dec}$ layers, respectively, and the output of each layer $l$ is specified as:
\begin{equation*}
\mathbf{h^l} = \texttt{tanh}(\mathbf{W^l}\cdot \mathbf{h^{l-1}} + \mathbf{b^l})
\end{equation*}
where the weight matrix at the $l^{th}$ layer, $\mathbf{W^l}$, is shared in the process of learning, $\mathbf{e_u}: u \in N$.

On the other hand, \texttt{embed} and \texttt{reconstruct} are a single-layer perceptron (SLP) defined as:
\begin{align}
\texttt{embed} =  \texttt{tanh}(\mathbf{W_{emb}} \cdot \mathbf{h^{L_{enc}}} + \mathbf{b_{emb}}) \nonumber \\
\texttt{reconstruct} = \texttt{tanh}(\mathbf{W_{rec}} \cdot \mathbf{h^{L_{dec}}} + \mathbf{b_{rec}}) \nonumber
\end{align}
where $\mathbf{W_{emb}}$ and $\mathbf{W_{rec}}$ are the weight matrices for the embedding and reconstruction layers, respectively,  which are also shared while learning the parameters $\mathbf{e_u}: u \in N$.
$\mathbf{h^{L_{enc}}}$ and $\mathbf{h^{L_{dec}}}$ are the activations of the last layers of the encoder and decoder, respectively.
$\mathbf{b^{l}}, \mathbf{b_{emb}}$ and $ \mathbf{b_{rec}}$ are the biases for the $l^{th}$-layer, embedding layer and reconstruction layer, in their respective order.

\subsubsection{Optimization} 
The model parameters are trained by minimizing the reconstruction errors on the weight and feature matrices $\mathbf{M}$ and $\mathbf{D}$, respectively as follows:
\begin{align}
\mathcal{L_C}^u &= \vert \vert (\mathbf{A_u} - \mathbf{\tilde{A}_u}) \odot \mathbf{s} + (\mathbf{D_u} - \mathbf{\tilde{D_u}}) \odot \mathbf{s} \vert \vert^2_2  \nonumber   \\
\mathcal{L_O}^u &= \vert \vert (\mathbf{M_u} - \mathbf{\tilde{M}_u}) \odot \mathbf{s} \vert \vert^2_2   \nonumber \\
\mathcal{L_I}^u &= \vert \vert (\mathbf{M_u}^T - \mathbf{\tilde{M}_u}^T) \odot \mathbf{s} \vert \vert^2_2  \nonumber  \\
\mathcal{L} &= \min_{\Theta} \sum_{u \in N} \mathcal{L_C}^u + \mathcal{L_O}^u + \mathcal{L_I}^u  \nonumber 
\end{align}
where $\odot$ is an element-wise multiplication, $\mathbf{A} = \mathbf{M} | \mathbf{M}^T$ is the undirected adjacency matrix, $\Theta$ denotes the set of all learnable model parameters, and $\mathbf{s}$ is a simple weighting trick adopted from~\cite{Wang:2016:SDN:2939672.2939753} to avoid the trivial solution of reconstructing the zeros for sparse matrices. 
That is, for the $j^{th}$ entry, $\mathbf{s_j} = \mu > 1$ if $\mathbf{A_{u,j}} \ge 1$ otherwise $\mathbf{s_j} = 1$ .
In all the experiments we use $\mu=10$.

Therefore, $\mathcal{L_C}$ enables us to preserve the combined neighborhood, in and out, and the content similarity between nodes.
$\mathcal{L_O}$ and $\mathcal{L_I}$ enables us to preserve the outgoing and incoming neighborhood of nodes, respectively.

\subsection{Edge Model} 
The limit of \dnode is that it does not satisfy the constraint that we specify in Remark~\ref{rem:proximity_remark}.
To solve this, we propose \dedge that has the same architecture depicted in Fig.~\ref{fig:diagram}, but iterates over the edges.
For each directed edge $(u,v) \in E$, we execute \diagram on both $u$ and $v$; we use however a simple trick by adjusting $\mathcal{L_I}$ for $u$ as: 
\[
\mathcal{L_I}^u = \vert \vert (\mathbf{M_v}^T - \mathbf{\tilde{M}_u}) \odot \mathbf{s} \vert \vert^2_2
\] 
In other words, the reconstructions of $u$'s out-neighborhood, $\mathbf{\tilde{M}_u}$, is ensured to be consistent with $v$'s actual incoming neighborhood, $\mathbf{M_v}^T$, consequently reducing the distance $\mathbf{o_u} - \mathbf{i_v}$ between $u$'s outgoing and $v$'s incoming embedding, and hence projecting them close to each other.

Although iterating over the edges in large networks might be expensive, we employ \textit{transfer learning} and use the trained parameters of the node model as a starting point.
In our experiments, transfer learning ensures convergence in one or two epochs, otherwise \dedge requires at least 30 epochs.

For both \dnode and \dedge we use dropout regularization as our datasets are extremely sparse.
\section{Experimental Evaluation}
\label{sec:experiments}
\begin{table}[b!]
\centering
\begin{tabular}{l|r|r|r|r}
\hline
  Dataset & $\vert N \vert$ & $\vert E \vert$ & $ d = \vert V \vert$ & \#Labels \\
  \hline
  Cora & 2,708  & 5,278 & 3,703 & 7 \\ \hline
  Citeseer & 3,312 & 4,660 & 1,433 & 6 \\ \hline
\end{tabular}
\vspace{5mm}
\caption{Dataset Summary}
\label{tbl:dataset_summary}
\end{table}

We have evaluated the performance of our algorithm against two popular citation network datasets summarized in Table~\ref{tbl:dataset_summary}, comparing it against the following state-of-the-art baselines.

\subsection{Baselines} 
We include six baselines that preserve both symmetric and asymmetric neighborhoods, as well as content similarity.
\begin{enumerate}
\item \hope~\cite{Ou:2016:ATP:2939672.2939751} is a direction-aware algorithm that uses a generalized variant of Singular Value Decomposition.
\item \deepwalk~\cite{Perozzi:2014:DOL:2623330.2623732} and \ntov~\cite{Grover:2016:NSF:2939672.2939754} are algorithms that learn a symmetric neighborhood of nodes using  \sg~\cite{Mikolov:2013:DRW:2999792.2999959} by leveraging sampled random walks.
\item \dane~\cite{ijcai2018-467}, \anrl~\cite{ijcai2018-438} and \vgae~\cite{kipf2016variational} preserve both neighborhood and content similarity of nodes, yet only in a symmetric manner.
\dane and \anrl use different architectures of autoencoders, whereas \vgae uses a variational autoencoder.
\end{enumerate}

We use three kinds of common network analysis tasks, which are \emph{network reconstruction}, \emph{link prediction}, and \emph{node classification} in order to evaluate the performance of the embedding algorithms.
\subsection{Settings}
The parameters of \diagram are tuned using random search, while for the baselines we use the source code provided by the authors and the optimal values reported in the corresponding papers.
For both datasets, \diagram's variants have the same configuration. 
The encoder layer configuration is $[\vert V \vert, 512, 256]$, and the embedding size $k$ is $128$.
The decoder layer configuration is $[128, 256, 512]$, and for reconstruction layer the configuration is $[512, \vert V \vert +  \vert F \vert ]$.
Dropout rate is equal to 0.2 (Cora) and 0.1 (Citeseer); the learning rate is 0.0001 for both.

\subsection{Network Reconstruction}
\label{sub:sec:network_reconstruction}

A good embedding algorithm should be able to preserve the structure of the network, and the goodness is usually verified by evaluating its performance in reconstructing the network.
Following the practice of related studies, we quantify the performance using the precision-at-k (P@K) metric.

To this end, we first compute the pairwise proximity between all nodes. 
For every pair of nodes $u$ and $v$, a direction-aware algorithm computes proximity $\texttt{prx}(u, v)$ between $u$ and $v$ as 
\[
\texttt{prx}(u, v) = \frac{1}{1 + e^{-\texttt{dot}(\mathbf{o_u}, \mathbf{i_v})}}
\]
and for direction-oblivious algorithms we simply have a single embedding $\mathbf{z_u}$ and $\mathbf{z_v}$ for $u$ and $v$, respectively, and hence $\texttt{prx}$ is computed as:
\[
\texttt{prx}(u, v) = \frac{1}{1 + e^{-\texttt{dot}(\mathbf{z_u}, \mathbf{z_v})}}
\]
Finally, node pairs are ranked according to their proximity score, and P@K is simply the precision measured with respect to the ground truth edges at cut-off $K$.

\begin{table}[t!]
\centering
\begin{tabular}{|l|l|l|l|l|l|}
\hline
\multirow{2}{*}{\textbf{Dataset}} & \multirow{2}{*}{\textbf{Algorithms}} & \multicolumn{4}{c|}{\textbf{P@K(\%)}} \\ \cline{3-6} 
 &  & \textbf{K=2500} & \textbf{K=5000} & \textbf{K=7500} & \textbf{K=10000} \\ \hline
\multicolumn{1}{|c|}{\multirow{8}{*}{\textbf{Citeseer}}} & \dedge & \textbf{57} & \textbf{44} & \textbf{37} & \textbf{31} \\ \cline{2-6} 
\multicolumn{1}{|c|}{} & \dnode & 46 & 36 & 29 & 25 \\ \cline{2-6} 
\multicolumn{1}{|c|}{} & \dane & 3 & 11 & 13 & 12 \\ \cline{2-6} 
\multicolumn{1}{|c|}{} & \anrl & 3 & 2 & 2 & 2 \\ \cline{2-6} 
\multicolumn{1}{|c|}{} & \vgae & 16 & 13 & 12 & 11 \\ \cline{2-6} 
\multicolumn{1}{|c|}{} & \hope & 46 & 38 & 30 & 24 \\ \cline{2-6} 
\multicolumn{1}{|c|}{} & \deepwalk & 5 & 5 & 4 & 4 \\ \cline{2-6} 
\multicolumn{1}{|c|}{} & \ntov & 19 & 16 & 16 & 12 \\ \hline
\multirow{8}{*}{\textbf{Cora}} & \dedge & \textbf{59} & \textbf{49} & \textbf{42} & \textbf{36} \\ \cline{2-6} 
 & \dnode & 46 & 38 & 31 & 30 \\ \cline{2-6} 
 & \dane & 10 & 10 & 9 & 0 \\ \cline{2-6} 
 & \anrl & 3 & 3 & 2 & 2 \\ \cline{2-6} 
 & \vgae & 21 & 18 & 15 & 14 \\ \cline{2-6} 
 & \hope & 48 & 42 & 36 & 30 \\ \cline{2-6} 
 & \deepwalk & 9 & 9 & 9 & 8 \\ \cline{2-6} 
 & \ntov & 6 & 6 & 7 & 10 \\ \hline
\end{tabular}
\vspace{5mm}
\caption{P@K results of the network reconstruction experiment}
\label{tbl:network_reconstruction_result}
\end{table}

The results of the network reconstruction experiments are reported in Table~\ref{tbl:network_reconstruction_result}; our algorithms--\dedge in particular--outperforms all the baselines with a significant margin.
In addition, as a consequence of satisfying Remark~\ref{rem:proximity_remark}, \dedge performs much better than \dnode.

The Cora dataset for instance has $\approx 5000$ edges, and we see that for $K=5000$, out of the $\approx 7,000,000$ possible pairs \dedge correctly filters out almost half of the ground truth edges.
These results are $7\%$ and $11\%$ better than \hope and \dnode, respectively,  and they are in stark contrast to the direction-oblivious baselines.


\subsection{Link Prediction}
\begin{figure}[t!]
\centering
\includegraphics[scale=0.5]{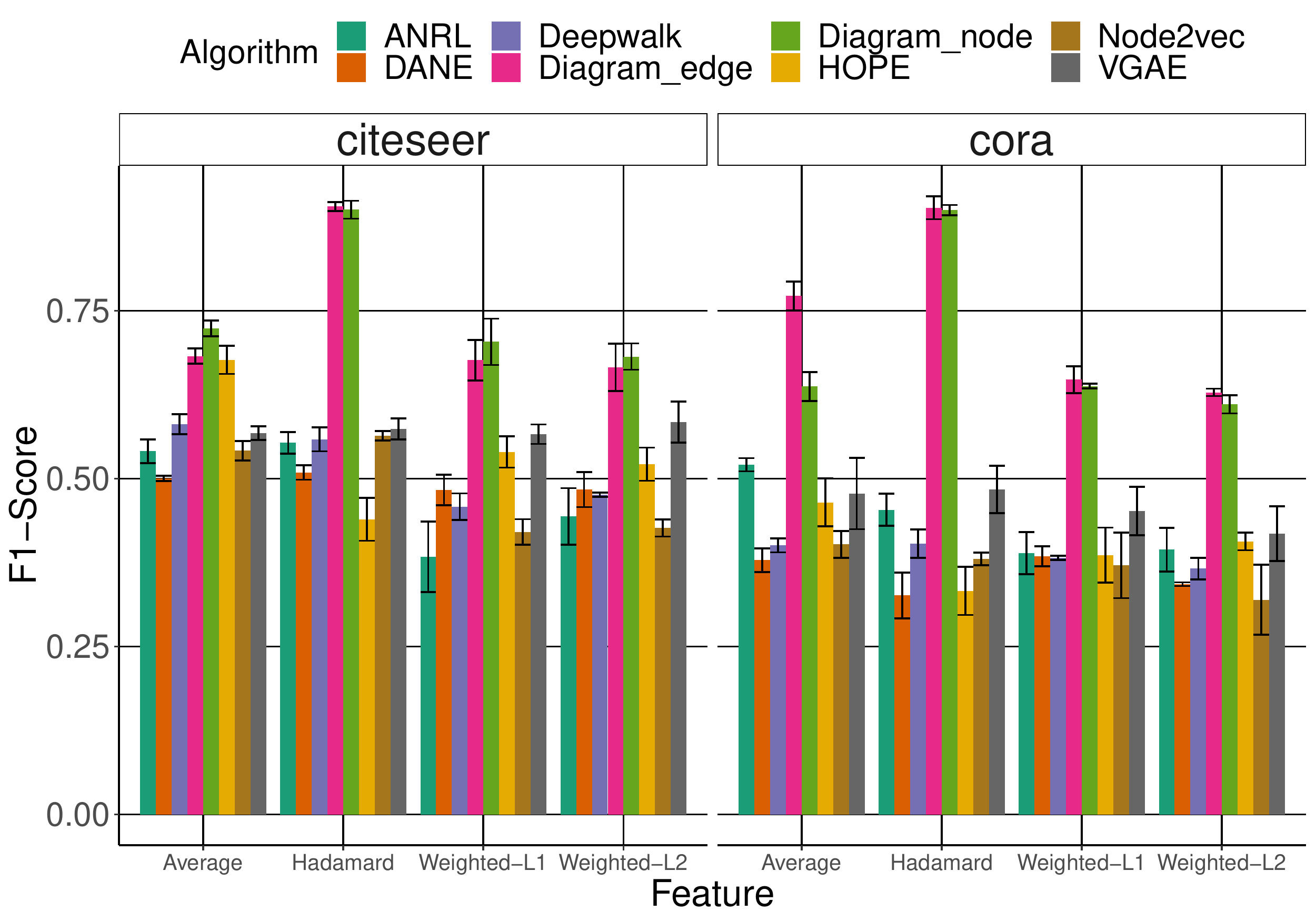}
\caption{F1-Score results for link prediction, and $p=10$.}
\label{fig:link_prediction}
\end{figure}

One of the most common real-world applications of graph embedding algorithms is link prediction. 
In this experiment, similar to~\cite{Grover:2016:NSF:2939672.2939754,Wang:2016:SDN:2939672.2939753,SKM18} we first sample $p$ percent of the edges from the graph. 
We refer to these edges as \emph{true samples} and remove them from the graph by ensuring that the residual graph remains connected. 
We also sample $p$ percent of pair of nodes that are not connected by an edge, referred to as \emph{false samples}.

To show how robust our approach is, we train the variants of \diagram on the residual graph, while the baselines are trained on the complete graph.
Then we use the learned embeddings of the algorithms to predict edges as follows.

For all the sampled edges, true and false, we construct edge features using the techniques proposed by~\cite{Grover:2016:NSF:2939672.2939754}, which are the average, element wise multiplication, weighted-L1, and weighted-L2 (Average, Hadamard, W-L1, W-L2, respectively) of the embeddings of the source and target nodes of each edge.
For direction-aware algorithms (\diagram and \hope), edge features are asymmetric, $\mathbf{o}$ and $\mathbf{i}$ are used, while the other baselines are symmetric, only $\mathbf{z}$ is used.
Then, for each sample, we assign the label `1' if it comes from the true ones, `0' otherwise.

We finally train a binary classifier using a 3-fold cross validation and  we report the mean of the area under the curve (AUC) score in Fig.~\ref{fig:link_prediction}, along with the error margin. 
\label{sub:sec:link_prediction}
Although \diagram's variants have been trained on the residual graph, yet we see that they significantly outperform all the baselines across all feature construction techniques.
This is due to the fact that \diagram uses direction-aware embeddings.
Even though \hope uses such kind of embeddings as well, its linear model is not powerful enough to capture the highly non-linear structure of real graphs~\cite{Wang:2016:SDN:2939672.2939753}. 
Furthermore, \dedge performs better than \dnode, mostly with smaller variance, with exception of W-L1.

\subsection{Node Classification}
\begin{table*}[t!]
\centering
\begin{tabular}{|l|l|l|l|l|l|}
\hline
\multirow{3}{*}{\textbf{Training Ration}} & \multirow{3}{*}{\textbf{Algorithm}} & \multicolumn{4}{c|}{\textbf{Dataset}} \\ \cline{3-6} 
 &  & \multicolumn{2}{c|}{\textbf{Cora}} & \multicolumn{2}{c|}{\textbf{Citeseer}} \\ \cline{3-6} 
 &  & \textbf{Micro-F1(\%)} & \textbf{Macro-f1(\%)} & \textbf{Micro-F1(\%)} & \textbf{Macro-F1(\%)} \\ \hline
\multirow{8}{*}{\textbf{10\%}} & \dedge & 75 & 73 & 60 & 56 \\ \cline{2-6} 
 & \dnode & 74 & 72 & 60 & 56 \\ \cline{2-6} 
 & \dane & \textbf{77} & \textbf{75} & 61 & 57 \\ \cline{2-6} 
 & \anrl & 74 & 72 & \textbf{66} & \textbf{62} \\ \cline{2-6} 
 & \vgae & 75 & 72 & 58 & 52 \\ \cline{2-6} 
 & \hope & 63 & 61 & 41 & 37 \\ \cline{2-6} 
 & \deepwalk & 70 & 68 & 48 & 45 \\ \cline{2-6} 
 & \ntov & 73 & 72 & 51 & 47 \\ \hline
\multirow{8}{*}{\textbf{30\%}} & \dedge & \textbf{80} & \textbf{79} & 68 & 64 \\ \cline{2-6} 
 & \dnode & 79 & 78 & 68 & 63 \\ \cline{2-6} 
 & \dane & 79 & 77 & 66 & 62 \\ \cline{2-6} 
 & \anrl & 77 & 75 & \textbf{71} & \textbf{67} \\ \cline{2-6} 
 & \vgae & 77 & 74 & 60 & 53 \\ \cline{2-6} 
 & \hope & 73 & 71 & 49 & 45 \\ \cline{2-6} 
 & \deepwalk & 77 & 76 & 55 & 51 \\ \cline{2-6} 
 & \ntov & 79 & 78 & 58 & 54 \\ \hline
\multirow{8}{*}{\textbf{50\%}} & \dedge & \textbf{82} & \textbf{80} & 70 & 65 \\ \cline{2-6} 
 & \dnode & 81 & 79 & 69 & 64 \\ \cline{2-6} 
 & \dane & 81 & 79 & 70 & 65 \\ \cline{2-6} 
 & \anrl & 78 & 76 & \textbf{73} & \textbf{68} \\ \cline{2-6} 
 & \vgae & 78 & 75 & 60 & 53 \\ \cline{2-6} 
 & \hope & 76 & 75 & 50 & 46 \\ \cline{2-6} 
 & \deepwalk & 80 & 79 & 56 & 52 \\ \cline{2-6} 
 & \ntov & 81 & \textbf{80} & 60 & 55 \\ \hline
\end{tabular}
\vspace{5mm}
\caption{F1-Micro and F1-Macro results for node classification experiments on Cora and Citeseer datasets. 
Each experiment is carried out using a fraction (10, 30, and 50 percent) of the data as training set and the rest as test set.}
\label{tbl:node_classification}
\end{table*}
Following the most common practices in these kinds of experiments, we train a multi-class one-vs-rest classifier using logistic regression.
We use learned embeddings as node features to predict the corresponding label.
Yet again, we perform a 10-fold cross-validation experiment; Table~\ref{tbl:node_classification} reports the average Micro-F1 and Macro-F1 metrics.

For Cora, in most cases \diagram performs better than \dane, \anrl, and \vgae, methods known for their superior performance in node classification.
For Citeseer, however, \diagram performs worst than \anrl, comparable to \dane and better than \vgae.

Note that \diagram is significantly better than \hope, the direction-aware baseline.
This is due to the fact that \diagram attempts to balance both direction and content by incorporating textual features and the undirected neighborhood.

\section{Related Work}
\label{sec:related_work}
Classical graph embedding techniques rely on matrix factorization techniques.
Fairly recently, however, several studies have been proposed based on shallow and deep neural networks~\cite{Pan:2016:TDN:3060832.3060886,ijcai2018-467,Wang:2016:SDN:2939672.2939753,ijcai2018-438,Perozzi:2014:DOL:2623330.2623732,Grover:2016:NSF:2939672.2939754,Yang:2015:NRL:2832415.2832542,SKM18}.
The earlier ones, such as \deepwalk~\cite{Perozzi:2014:DOL:2623330.2623732} and \ntov~\cite{Grover:2016:NSF:2939672.2939754}, are techniques based on sampled random walks that are intended to capture the local neighborhood of nodes. 
Virtually all methods based on random walks rely on the \sg model~\cite{Mikolov:2013:DRW:2999792.2999959}, whose objective is to maximize the probability of a certain neighbor node $v \in N$ of an anchor node $u \in N$, given the current embedding $\mathbf{z_u}$ of $u$.
One of the main drawbacks of random walk based methods is their sampling complexity.

To address this limitation, several follow up studies, e.g.~\cite{ijcai2018-467,Wang:2016:SDN:2939672.2939753,ijcai2018-438}, have been proposed based on deep-feed forward neural networks, such as Deep Autoencoders.
However, all these methods are inherently applicable for undirected networks, as they do not explicitly care for directionality.

A method called Hope~\cite{Ou:2016:ATP:2939672.2939751} has been proposed for preserving important properties in directed networks.
However, it relies on matrix factorization, and do not effectively capture the highly non-linear structure of the network~\cite{Wang:2016:SDN:2939672.2939753}.

Our algorithm addresses the aforementioned limitations, however the results of the node-classification experiment are not satisfactory and we intend to carefully investigate the problem and design a better algorithm in the complete version of this paper.
\section{Conclusion}
\label{sec:conclusion}

In this study we propose an ongoing, yet novel direction-aware, text-enhanced graph embedding algorithm called \diagram.
Unlike most existing studies that trade-off between learning high-quality embeddings for either link prediction or node classification, \diagram is capable to balance both.

We have empirically shown \diagram's superior performance in asymmetric link prediction and network reconstruction and comparable performance in node classification over the state-of-the-art.
However, we are still investigating directions to improve \diagram so as to improve its performance on node classification as well and we also seek to incorporate more datasets.
We shall cover all of these aspects in a future version of this study.

\clearpage

\bibliography{acl2019}
\bibliographystyle{IEEEtran}


\end{document}